\documentclass{article}

    \usepackage[preprint]{score_workshop}

\usepackage[utf8]{inputenc} 
\usepackage[T1]{fontenc}    
\usepackage{hyperref}       
\usepackage{url}            
\usepackage{booktabs}       
\usepackage{amsfonts}       
\usepackage{nicefrac}       
\usepackage{microtype}      
\usepackage{xcolor}         
\usepackage{float}

\usepackage{natbib}
\setcitestyle{numbers, open={[}, close={]}}
\usepackage{hyperref}
\hypersetup{colorlinks, citecolor=orange, linkcolor=red, urlcolor=blue}
\usepackage{listings}
\usepackage{pythonhighlight}
\usepackage{bm}

\usepackage[ruled]{algorithm2e}

\usepackage{amsmath}

\DeclareMathOperator*{\argmin}{arg\,min}
\DeclareMathOperator*{\E}{\mathbb{E}}

\usepackage{graphicx}
\usepackage{wrapfig}

\usepackage{enumitem}
\usepackage{amsthm}

\def\boldz{\mathbf{z}}
\def\boldx{\mathbf{x}}
\def\bolds{\mathbf{s}}
\def\bolda{\mathbf{a}}
\def\boldepsilon{\bm{\epsilon}}

\def\boldtheta{\bm{\theta}}
\def\boldc{\mathbf{c}}

\title{
    Conditional Diffusion with Less Explicit Guidance \newline
    via Model Predictive Control
}

\author{%
    Max W. Shen\thanks{Corresponding author: shen.max@gene.com} ,
    Ehsan Hajiramezanali, \\
    \textbf{
        Gabriele Scalia,
        Alex Tseng,
        Nathaniel Diamant,
        Tommaso Biancalani,
        Andreas Loukas
    }
    \\
    Genentech\\
    \texttt{shen.max@gene.com}
}

\begin{document}

\maketitle

\begin{abstract}
    How much explicit guidance is necessary for conditional diffusion?
    We consider the problem of conditional sampling using an unconditional diffusion model and limited explicit guidance (e.g., a noised classifier, or a conditional diffusion model) that is restricted to a small number of time steps.
    We explore a model predictive control (MPC)-like approach to approximate guidance by simulating unconditional diffusion forward, and backpropagating explicit guidance feedback.
    MPC-approximated guides have high cosine similarity to real guides, even over large simulation distances.
    Adding MPC steps improves generative quality when explicit guidance is limited to five time steps.
\end{abstract}

\vspace{-.1 in}
\begin{figure}[h]
    \centering
    \includegraphics[width=0.5\textwidth]{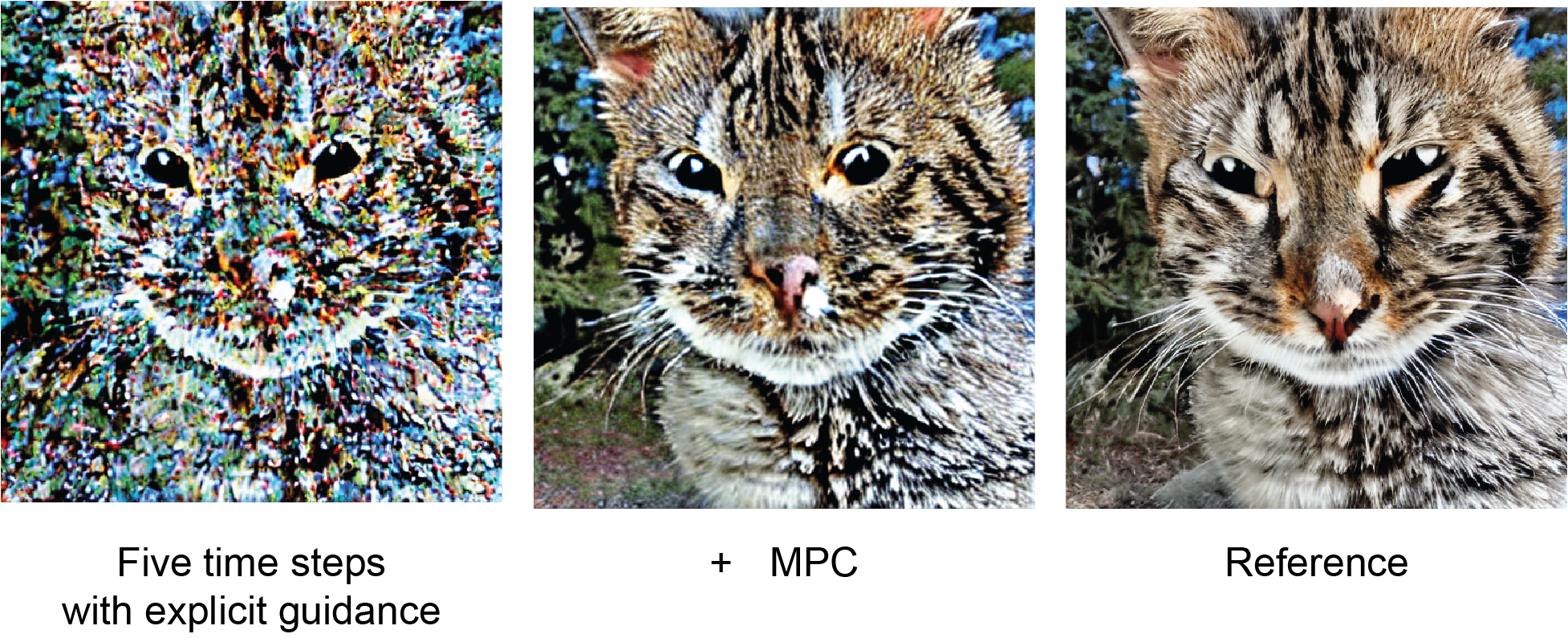}
\end{figure}


\vspace{-0.1 in}
\section{Introduction}
\vspace{-0.1 in}

Diffusion models are a class of generative models that have achieved remarkable sample quality, particularly for text-to-image generation \cite{stablediffusion}, where diffusion has been guided using \textit{classifier guidance} or \textit{classifier-free guidance} to sample images $\boldx \sim p(\boldx | \boldc)$ for a conditioning variable $\boldc$ (e.g., text) \cite{dhariwal, ho2021classifierfree}.
Controlling generative models is important for applications such as text generation and drug discovery, where multiple distinct conditional variables $\boldc_1, \boldc_2, ... \boldc_n$ can be important: e.g., drug activity and permeability \cite{lambo}.

For each new conditioning information source $\boldc$ of interest, 
classifier guidance and classifier-free guidance require training a new explicit guidance model over all diffusion time steps $t \in [0, ..., T]$ (often, $T=$100 to 1,000),
and sample using the explicit guide at each generative time step (often, 25-100) 
\cite{dhariwal, ho2021classifierfree}.
Here, we explore whether conditional sampling is achievable without explicit guidance at every generative step,
and if it is achievable with very few steps.
This line of inquiry may make it easier to condition on new variables by reducing the training burden of new explicit guidance models.

Rejection sampling and Langevin "churning" have been explored for image editing, inpainting, and conditional sampling on new variables without training a new model over diffusion time steps, but lack general applicability \citep{sdedit, stablediffusion, ilvr, singh, comecloserdiffusefaster, d2c, repaint}:
churning appears limited to "local" edits, while rejection sampling is inefficient for rare events.
Separately, scheduler advances have reduced sampling steps from 100-1000 to 25-50 while retaining high sample quality \citep{karras, liu2022pseudo}.
This work aims to be generally applicable and synergistic with scheduler improvements.
\paragraph{Diffusion models.}

Diffusion models are trained on noise-corrupted data, and learn an iterative denoising process to generate samples.
We give a non-precise introduction following \cite{imagen}, and refer interested readers to \cite{karras} for a precise description.
A diffusion model $\hat{\boldx}_\theta$ is trained to optimize:

\vspace{-0.1 in}
\begin{equation}
    \E_{\boldx, \boldc, \boldsymbol{\epsilon}, t} [
        w_t \|
            \hat{\boldx}_\theta (\alpha_t \boldx + \sigma_t \boldsymbol{\epsilon}, \boldc) - \boldx
        \|_2^2 
    ] 
\end{equation}
\vspace{-0.1 in}

where $(\boldx, \boldc)$ are data-conditioning pairs, $t \sim \mathcal{U}([0, 1])$, $\boldepsilon \sim \mathcal{N}(\mathbf{0, I})$, and $\alpha_t, \sigma_t$, and $w_t$ are time-varying weights that influence sample quality.
In the $\boldepsilon$-prediction parameterization, 
$\hat{\boldx}_\theta (\boldz_t, \boldc) = (\boldz_t - \sigma_t \boldepsilon_{\theta}(\boldz_t, \boldc)) / \alpha_t$ where $\boldepsilon_\theta$ is the learned function. 
Notably, this training procedure has an expectation over $t$, which can be hundreds to thousands of time steps. 

To sample, a simple scheduler starts at $\boldz_T \sim \mathcal{N}(\mathbf{0, I})$ and iteratively generates 
$\boldz_{t-1} = (\boldz_t - \sigma \tilde{\boldepsilon}_\theta) / \alpha_t$
where the choice of $\tilde{\boldepsilon}_\theta$ distinguishes sampling strategies.
In general, schedulers can jump to $\boldz_{t-\Delta}$ as a function of starting time $t$, jump size $\Delta$, latent $\boldz_t$, and predicted noise $\tilde{\boldepsilon}_\theta$.

\vspace{-0.1 in}
\paragraph{Diffusion guidance.}
Classifier guidance \citep{dhariwal} requires training a \textit{noised classifier} $p_t(\boldc | \boldz_t)$ over $T$ time steps, and uses 
$\tilde{\boldepsilon}_\theta = \boldepsilon(\boldz_t, \boldc) - \nabla_{\boldz_t} \log p_t(\boldc | \boldz_t)$.
Notably, pre-trained \textit{clean-data classifiers} cannot be directly used for guidance.
Classifier-free guidance \citep{ho2021classifierfree} learns both a conditional and unconditional diffusion model by setting $\boldc = \mathbf{0}$ with $10\%$ probability during training; 
$\tilde{\boldepsilon}_\theta = \boldepsilon_\theta(\boldz_t) := \boldepsilon_\theta(\boldz_t, \boldc = \mathbf{0})$ achieves unconditional sampling.
Classifier-free guidance with weight $w$ uses

\vspace{-0.1 in}
\begin{equation}
    \label{eq:cfg}
    \tilde{\boldepsilon}_\theta = (1+w) \boldepsilon(\boldz_t, \boldc) - w \boldepsilon_{\boldtheta}(\boldz_t).
\end{equation}

\vspace{-0.1 in}
\paragraph{Model predictive control (MPC).}

Model predictive control aims at controlling a time-evolving system in an optimized manner, by using a predictive dynamics model of the system and solving an optimization problem online to obtain a sequence of \textit{control actions}.
Typically, the first control action is applied at the current time, then the optimization problem is solved again to act at the next time step \cite{diffmpc}. The general formalized MPC problem is:

\vspace{-0.1 in}
\begin{equation}
    \argmin_{\bolds_{1:T}, \bolda_{1:T}}
    \sum_{t=1}^T \ell_t(\bolds_t, \bolda_t)
    \textrm{   subject to   } 
    \bolds_{t+1} = f(\bolds_t, \bolda_t); \bolds_1 = \bolds_{\texttt{init}}
\end{equation}
\vspace{-0.1 in}

where $\bolds_t, \bolda_t$ are the state and control action at time $t$, $\ell_t$ is a cost function, $f$ is a dynamics model, and $\bolds_{\texttt{init}}$ is the initial state of the system. 
MPC can be solved with gradient methods \cite{neuralmpc, pmlr-v120-bharadhwaj20a}.




\vspace{-0.1 in}
\section{Approximate conditional guidance via model predictive control}
\vspace{-0.1 in}

\begin{wrapfigure}{l}{0.4\textwidth}
    \vspace{-0.3 in}
    \centering
    \includegraphics[width=0.4\textwidth]{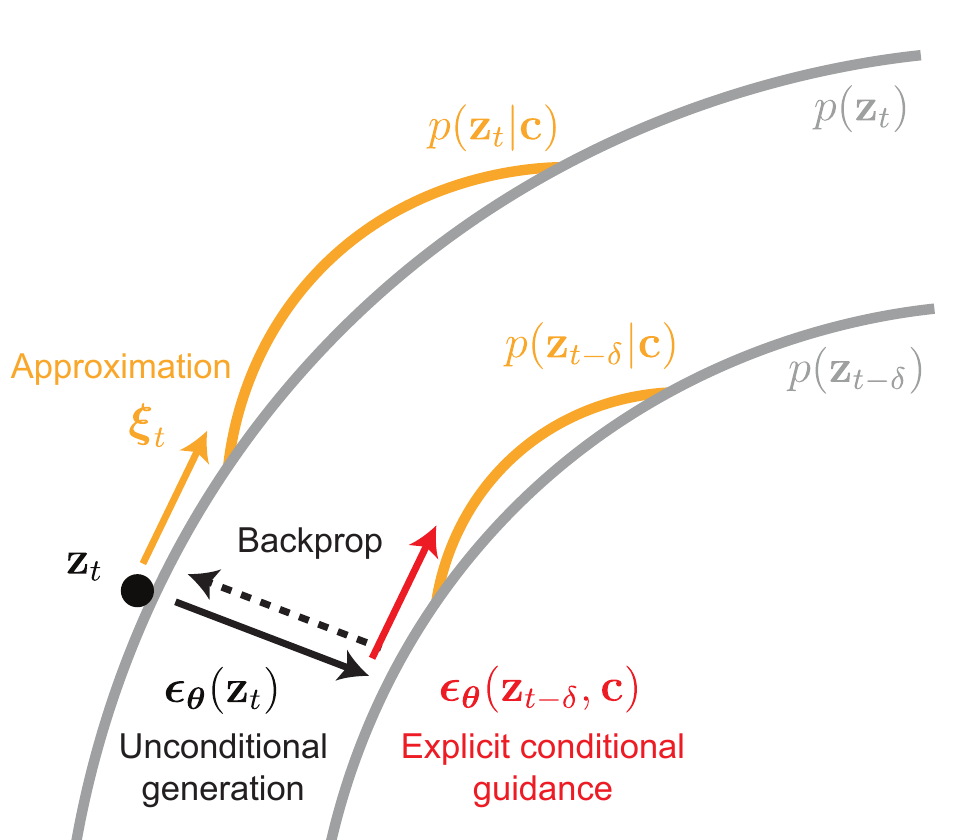}
    \label{fig:overview}
    \vspace{-0.4 in}
\end{wrapfigure}

Our problem is performing conditional diffusion on a latent $\boldz_t$
with only access to an unconditional diffusion model.
In particular, we do not have an explicit conditional guide $\boldepsilon_{\boldtheta} (\boldz_{t}, \boldc)$ at time $t$;
instead, we can evaluate it only at $t-\delta$.
Our method, MPC guidance, optimizes an approximation
$\boldsymbol{\xi}_t \approx \boldepsilon_{\boldtheta}(\boldz_t, \boldc)$,
which is used in classifier-free guidance (eq. \ref{eq:cfg}) to apply one generative step on $\boldz_t$ to obtain $\boldz_{t-\Delta}$.
This can be applied repeatedly to reach $\boldz_0$.
In terms of MPC, we view $\boldz_t$ as states, control actions as $\tilde{\boldepsilon}_\theta$, the dynamics model $f$ as the diffusion generative process given $\boldz_t$ and $\tilde{\boldepsilon}_\theta$,
and define loss $\ell$ at time $t-\delta$ using the explicit guide (Fig. \ref{fig:overview}).



\paragraph{Noised classifier.} 
With a noised classifier $p_{t-\delta}(\boldc|\boldz_t)$, the explicit guide
$\boldepsilon_{\boldtheta} (\boldz_{t-\delta}, \boldc) =
\nabla_{\boldz_{t-\delta}} \log p_{t-\delta}( \boldc | \boldz_{t-\delta})$. 
We propose to unconditionally generate $\boldz_{t-\delta}$ from $\boldz_t$ and evaluate $\log p_{t-\delta}(\boldc|\boldz_{t-\delta})$ which we treat as "inverse loss".
Our MPC guide at time $t$ is a first-order, one-step optimization of this loss:

\vspace{-.1 in}
\begin{equation}
    \boldsymbol{\xi}_t = - \nabla_{\boldz_t} \ell (\boldz_{t-\delta}) = - \nabla_{\boldz_t} \log p_{t-\delta}(\boldc | \boldz_{t-\delta})
\end{equation}

\vspace{-.1 in}
\begin{algorithm}[htp]
    \caption{Approximate guide with noised classifier}
    \label{alg:noised}
    \SetAlgoLined
    \DontPrintSemicolon
    \small
\begin{verbatim}
def approx_guide(zt, t, dt, noised_classifier):
    z = denoise(zt, t, dt)    # differentiable; denoise zt to time t-dt
    return autograd(noised_classifier(z), zt)   # grad wrt zt
\end{verbatim}

\end{algorithm}
\vspace{-.1 in}

\paragraph{Conditional diffusion model.} 
When the explicit guide is a conditional diffusion model $\boldepsilon_\theta(\boldz_{t-\delta}, \boldc)$, we denoise 
$\boldz_t$ to $\boldz_{t-\delta}$ 
and construct the MPC guide as:

\vspace{-.1 in}
\begin{equation}
    \boldsymbol{\xi}_t = - \nabla_{\boldz_t} \ell (\boldz_{t-\delta}) = - \nabla_{\boldz_t} \| \boldz_{t-\delta} - \boldz^* \| ^2
\end{equation}

where gradients with respect to $\boldz_t$ are blocked for the target
$\boldz^* := \boldz_{t-\delta} + \boldepsilon_{\boldtheta} (\boldz_{t-\delta}, \boldc )$.



\vspace{-.1 in}
\begin{algorithm}[htp]
    \caption{Approximate guide with conditional diffusion model}
    \label{alg:cond}
    \SetAlgoLined
    \DontPrintSemicolon
        \small
\begin{verbatim}
def approx_guide(zt, t, dt, cond_score):
    z = denoise(zt, t, dt)    # differentiable; denoise zt to time t-dt
    with no_grad():
        target = z + cond_score(z, t-dt)
    loss = (z - target)**2
    return autograd(loss, zt)   # grad wrt zt
\end{verbatim}

\end{algorithm}
\vspace{-.1 in}

\paragraph{Backpropagation through diffusion.} 
To compute gradients with respect to $\boldz_t$, we must backpropagate through unconditional diffusion.
This incurs memory cost linear in the number of denoising steps used.
In practice, five to ten denoising steps enabled good performance without memory issues.

\vspace{-0.1 in}
\section{Experiments}
\vspace{-0.1 in}

We perform experiments on Stable Diffusion \cite{stablediffusion}, an open-source text-to-image latent diffusion model trained on LAION-5B \cite{schuhmann2022laionb} with a pre-trained text conditional and unconditional model.
Latent diffusion occurs over 1000 time steps: $\boldz_0 \rightleftharpoons \boldz_{1000}$, and an adversarially-trained autoencoder encodes and decodes $\boldx \rightleftharpoons \boldz_0$.
We treat the conditional diffusion model as the explicit guide.
We use the pseudo linear multi-step (PLMS) scheduler \cite{liu2022pseudo} which is deterministic.

\begin{figure}[h]
    \centering
    \includegraphics[width=0.4\textwidth]{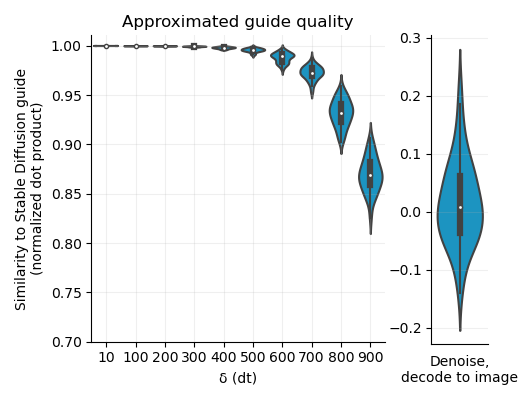}
    \caption{MPC guides have high cosine similarity to real guides}
    \label{fig:consistency}
\end{figure}

\vspace{-0.1 in}
\paragraph{Approximate guides have high accuracy.}
In figure \ref{fig:consistency}, we compare our approximated guide $\boldsymbol{\xi}_t$ to Stable Diffusion's conditional guide $\boldepsilon_{\boldtheta}(\boldz_t, \boldc)$ using the cosine similarity between the two gradients (see appendix for full details).
Approximate guides obtained by denoising $\boldz_t$ to $\boldz_{t-\delta}$ are very similar to Stable Diffusion's guide, with cosine similarity above 0.99 even as $\delta$ increases to 500 time steps out of 1000 total diffusion steps.
At $\delta=900$, similarity is maintained above 0.80.

In contrast, approximate guides formed by denoising and decoding $\boldz_t$ to images $\boldx$, applying CLIP \cite{clip} spherical loss, and backpropagating back to $\boldz_t$ are essentially orthogonal to $\boldepsilon_{\boldtheta}(\boldz_t, \boldc)$, with mean similarity around 0.01. 
This is consistent with observations that the manifold of natural images is complex in pixel space, and gradients on images are difficult to use for optimizing latents \cite{Plumerault2020Controlling}.
This highlights the challenge of conditional diffusion sampling using only clean-data classifiers.

\vspace{-0.1 in}
\paragraph{Approximate guides improve robustness to sample quality damage with reduced explicit guidance.}

We evaluated conditional sampling with explicit guidance restricted to just $n=5$ time steps, with classifier weight $w = 2$.
We compare to using $k=3$ additional MPC-guided generative steps (with a total of $n+k=8$ steps), and a \textit{reference} with full explicit guidance on $n+k$ steps - if MPC is accurate, then samples should look similar to the reference. 
We also generated \textit{gold standard} samples with 50 explicit guidance steps.
We evaluated PLMS baselines with both $n$ and $n+k$  generative steps (see appendix).
Each approach was initialized with identical $z_T$; as each approach is deterministic, quality can be judged by similarity to the reference and gold standard.

\begin{figure}[h]
    \centering
    \includegraphics[width=0.8\textwidth]{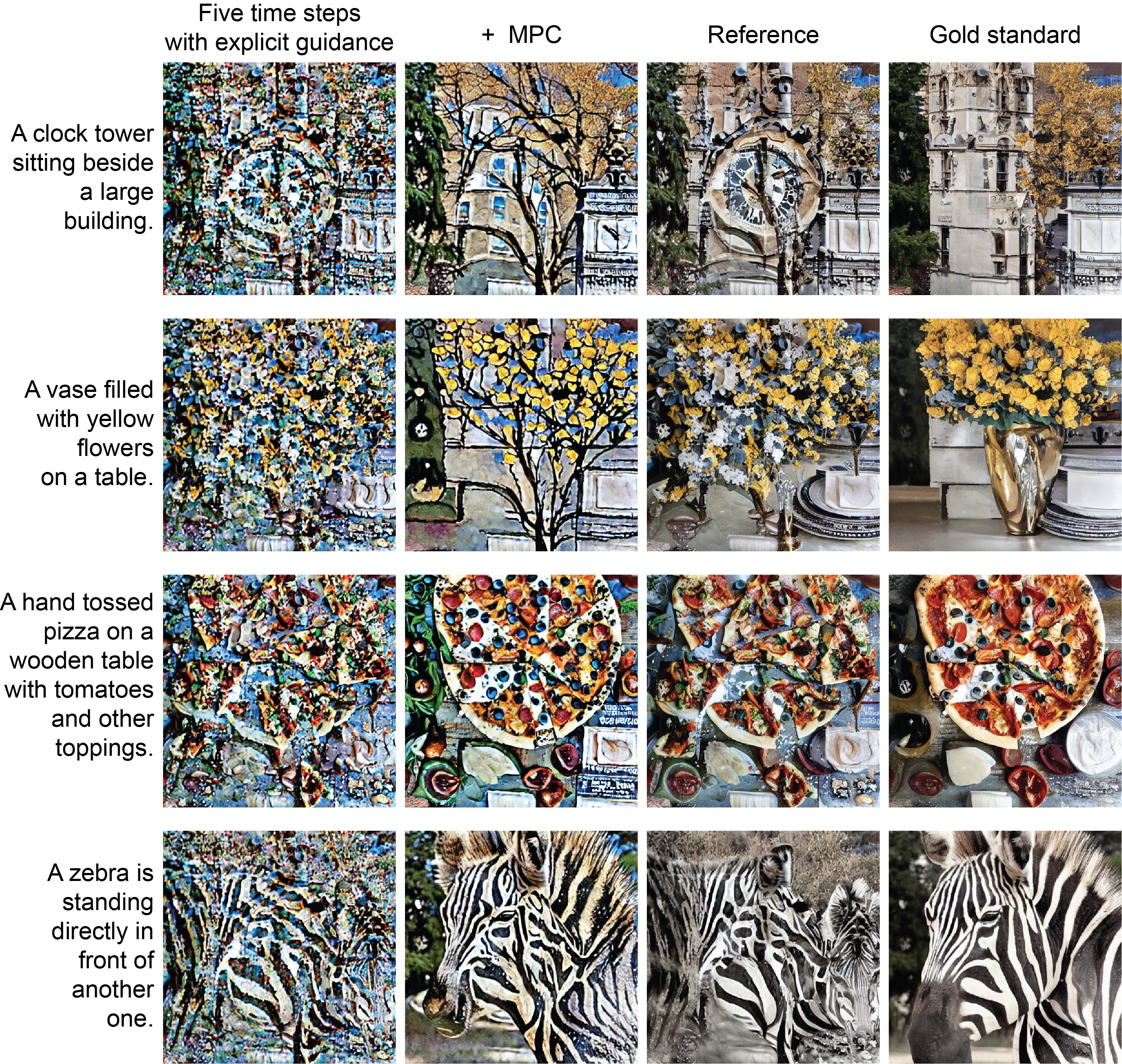}
    \caption{Comparison of samples (Stable Diffusion, pseudo linear multi-step scheduler, guidance weight $w=2$)}
    \label{fig:fig2}
\end{figure}

On random MS-COCO prompts, adding MPC generative steps significantly improved visual sample quality over the baseline (Fig. \ref{fig:fig2}) and improve FID to the reference and gold standard (Table \ref{table:fid}).
MPC samples are more visually similar to the reference than the baseline, and intriguingly, in some cases seem to outcompete the reference in visual similarity to the gold standard.


\vspace{-0.1 in}
\begin{table}[H]
    \centering
    \begin{tabular}{ l l l } \\ \toprule  
         FID ($\downarrow$) & Reference   & Gold standard \\
        \midrule
        Baseline ($n=5$)   & 400.0               & 443.28            \\
        + MPC ($k=3$)      & \textbf{282.4}      & \textbf{312.84}   \\
        \bottomrule
    \end{tabular}
    \label{table:fid}
\end{table}
\vspace{-0.1 in}

\vspace{-0.1 in}
\vspace{-0.1 in}


\vspace{-0.1 in}
\section{Discussion}
\vspace{-0.1 in}

We described a method for approximating guidance for conditionally sampling from diffusion models with model predictive control, and showed preliminary evidence that approximated guidance improves sample quality when access to a conditional guide is severely restricted to just five time steps.

Looking forward, future work may be interested in addressing instabilities and divergence.
In some settings, we found that approximate guides tended to cause divergence to reference latent trajectories over time.
We found this issue to be particularly problematic with larger classifier guidance weights $w$: 
even if $\boldsymbol{\xi}_t$ is very similar to $\boldepsilon(\boldz_t, \boldc)$ (e.g., 0.9999), and $\boldepsilon_{\boldtheta}(\boldz_t)$ is identical, the adjusted prediction
$\tilde{\boldepsilon}(\boldz_t, \boldc) = (1+w) \boldsymbol{\xi}_t - w \boldepsilon_{\boldtheta}(\boldz_t)$ 
can have significantly lower similarity (e.g., 0.992).
We also observed that divergence increased with the number of approximate guidance steps.

Our results suggest the possibility of conditional diffusion using explicit guidance (e.g., a conditional diffusion model) trained on a small number of time steps. 
Future work can explore this by restricting conditional training; here, we only restricted the time steps at which we queried the ground-truth guide which was trained on all time steps.



\newpage

\bibliography{biblio}



\newpage
\appendix

\section{Appendix}

\subsection{Experiments}

We used an Nvidia A100 with 80 GB memory for our experiments. 
Backpropagating through diffusion requires backpropagating through Stable Diffusion's U-Net several times.
We found that roughly 10 or more denoising steps exceeded the memory of our A100, but that five denoising steps was sufficient for performance.

We used classifier-free guidance weight $w=2$, following \cite{ho2021classifierfree}.
In practice, we scale our approximate guide 
$\boldsymbol{\xi}_t$ at time $t$ to match the norm of 
the unconditional score $\boldepsilon_{\boldtheta}(\boldz_{t})$.

We will release our code in a future version.

\paragraph{Details on Stable Diffusion.}
Stable Diffusion was trained with classifier-free guidance, conditioning on CLIP-embedded text prompts \cite{clip}, with $1000$ diffusion time steps.
An adversarially-trained autoencoder encodes and decodes images , which is an $8 \times$ down-sampled latent space. 
Latents $z_0$ were very weakly regularized ($10^{-6}$ weight) towards a unit Gaussian. 
Despite this, when visualized as images, latents $\boldz_0$ appear as fuzzy versions of the decoded image $\boldx = \texttt{decoder}(\boldz_0)$ \cite{stablecompression}. 

\paragraph{Similarity study.}
At each starting time $t$, we initialized $z_t$ by unconditionally denoising from the prior $z_T$.
We obtained an approximate guide for various $\delta$, also called $dt$.
Stable diffusion has 1000 total diffusion timesteps, so we varied $t$ in $[200, 400, 600, 800, 1000]$. 
We varied $\delta$ in increments of $100$, and performed 10 replicates for each experimental condition.
We used the following text prompts, some of which were from the Stable Diffusion paper \cite{stablediffusion}: 'a photo of a cat', 'a photo of an astronaut riding a horse on mars', 'a street sign that reads latent diffusion', 'a zombie in the style of picasso', 'a watercolor painting of a chair that looks like an octopus', 'an illustration of a slightly conscious neural network'.
We observed similar results for all prompts.
The plot depicts data for $t=1000$, for varying $\delta$ on the x-axis, across prompts and replicates: there are 60 datapoints for each violin plot, which is smoothed with kernel density estimation using seaborn.

\paragraph{Restricted explicit guidance experiments.}
Our approach used an eight-step schedule evenly divided from $t=$1000 to 0: [875, 750, 625, 500, 375, 250, 125, 0], with explicit guidance at times [750, 500, 250, 125, 0] and MPC at [875, 625, 375].
We compare to a \textit{reference} with the same eight-step schedule with full explicit guidance.
Our PLMS \textit{baseline} uses the five-step schedule [800, 600, 400, 200, 0] with explicit guidance.
We also tried another baseline using the eight-step schedule, explicit guidance at five time steps, and unconditional steps at times [875, 625, 375], but found that this baseline ignored prompts.

\paragraph{Wall-clock time (for one sample).}
50 generative steps takes about 12 seconds.
10 generative steps takes about 2.5 seconds.
We find that with 5 generative steps, adding 3 MPC steps adds negligible runtime, with all runs finishing in 1-3 seconds.
In a separate unreported experimental setting, our method, with 25 total generative denoising steps, guidance at 10 time steps, 10 unconditional denoising steps for approximating the guide, and churning, takes about 34 seconds.
The same setting, without churning, takes about 18 seconds.

\end{document}